\documentclass{article}
\usepackage[utf8]{inputenc}
\usepackage{graphicx}

\usepackage{pgfplotstable}
\pgfplotsset{compat=1.11}

\usepackage{times}
\usepackage{natbib}
\usepackage{hyperref}

\begin{document}

\title{DDD17: End-To-End DAVIS Driving Dataset}

\author{
	Jonathan Binas,
	Daniel Neil,
	Shih-Chii Liu, and
	Tobi Delbruck\footnote{Corresponding author: tobi@ini.uzh.ch}. \\ \\
	Institute of Neuroinformatics, \\ University of Zurich and ETH Zurich, Switzerland}

\vskip 0.3in

\maketitle

\begin{abstract} 
 Event cameras, such as dynamic vision sensors (\textbf{DVS}), and dynamic and active-pixel vision sensors (\textbf{DAVIS}) can supplement other autonomous driving sensors by providing a concurrent stream of standard active pixel sensor (\textbf{APS}) images and DVS temporal contrast events.
 The APS stream is a sequence of standard grayscale global-shutter image sensor frames. The DVS events represent brightness changes occurring at a particular moment, with a jitter of about a millisecond under most lighting conditions.
 They have a dynamic range of {\textgreater}120\,dB and effective frame rates {\textgreater}1\,kHz at data rates comparable to 30\,fps (frames/second) image sensors. 
 To overcome some of the limitations of current image acquisition technology, we investigate in this work the use of the combined DVS and APS streams in end-to-end driving applications.
The dataset DDD17 accompanying this paper is the first open dataset of annotated DAVIS driving recordings.
 DDD17 has over 12\,h of a 346x260 pixel DAVIS sensor recording highway and city driving in daytime, evening, night, dry and wet weather conditions, along with vehicle speed, GPS position, driver steering, throttle, and brake captured from the car's on-board diagnostics interface.
 As an example application, we performed a preliminary end-to-end learning study of using a convolutional neural network that is trained to predict the instantaneous steering angle from DVS and APS visual data.
\end{abstract}

\section{Introduction}
The rapid improvement of machine learning and computer vision systems has spurred the development of self driving vehicles, which have already covered millions of kilometers in real world scenarios.
It appears that the development of processing technology and algorithms currently advances at greater speed than the development of sensing hardware for capturing the necessary information from the surroundings of the vehicle, such as obstacles, traffic, marks, and signs.
Automotive image sensors are being intensively developed to deal with the conflicting requirements for low cost, high dynamic range, high sensitivity, and resistance to artifacts from flickering light sources such as LED traffic signs and car taillights.
Operation under bad weather and/or lighting conditions is a primary requirement for automotive self driving or automatic driver assistance systems (ADAS), however, current ADAS sensors and systems still face many problems compared to human driver performance in challenging situations.
Since event cameras have been proposed as possible ADAS sensors \citep{posch_retinomorphic_2014}, we collected data to study the use of an event camera to augment conventional imager technology.

Rather than providing frame-based video as output, the event camera dynamic vision sensor (DVS) detects local changes in the brightness of individual pixels and asynchronously outputs those changes at the time of occurrence \citep{Lichtsteiner_etal08,posch_retinomorphic_2014}.
Thus, only the parts of the scene that change produce data, lowering the output data rate, increasing the temporal resolution and reducing the latency in comparison to frame-based systems, since changes in pixel brightness are streamed out of the camera as they occur.
The local instantaneous gain control increases usability under uncontrolled lighting conditions.
The higher temporal resolution and limited data rate makes the DVS  well suited for autonomous driving applications, where both latency and power consumption are important.
A dynamic and active-pixel vision sensor (DAVIS) has pixels that concurrently output DVS events and standard image sensor intensity frames \citep{Brandli_etal14}.

Recent studies have shown the utility of using DVS in data-driven convolutional neural network (CNN) real time applications 
\citep{moeys_steering_2016b,lungu_live_2017}. 
In these applications, DVS input frames typically consist of a 2D histogram image of a constant number of a few thousand DVS events.
Because the DVS event rate is proportional to the rate of change of brightness, i.e. scene reflectance \citep{Lichtsteiner_etal08}, the CNN frame rate is variable, ranging from about 1\,fps up to 1000\,fps. 
\citet{moeys_steering_2016b} showed that combining the standard image sensor frames from the sensor with the DVS frames resulted in higher accuracy and lower average reaction time.
Here we extend this work to real world driving in the first published end-to-end dataset of DVS or DAVIS driving data.

\section{Davis Driving Dataset 2017 (DDD17)}

DDD17 is available from \href{http://sensors.ini.uzh.ch/databases.html}{sensors.ini.uzh.ch/databases}. This data is collected from Swiss and German road driving under various conditions. It includes DAVIS data and car data.
Since the main aim of this dataset is to enable studying the fusion of APS and DVS data for ADAS, we did not include other sensors such as LIDAR.

\subsection{DAVIS data}
 Visual data was captured using a DAVIS346B prototype camera, containing a DAVIS APS+DVS camera, such that event-based and traditional frame-based data could be recorded at the same time, through the same optics.
 The camera resolution is $346 \times 260$ pixels. 
 The camera architecture is similar to \citet{Brandli_etal14},
 but the sensor has 2.1X more pixels and includes on-chip column parallel analog to digital converters (ADCs) for frame-based APS output up to 50\,fps.
 The DAVIS346B also has optimized buried photodiodes with microlenses that increase fill factor and reduce dark current, thereby improving operation at low light intensities by factor of about 4 compared with the \citet{Brandli_etal14} DAVIS240C.
 A fixed focal length lens (C-mount, 6mm) was used for all recordings, 
 providing a horizontal field of view of 56${^\circ}$.
 The aperture was set manually, depending on lighting conditions. 
 The APS frame rate depended on exposure duration 
 to a value between 10\,fps and 50\,fps; 
 in some recordings it varied depending on the auto-exposure duration algorithm.
 The frames were captured using the DAVIS global shutter mode to minimize motion artifacts.
 The camera was mounted using a glass suction tripod mount behind the windshield, 
 just below the rear mirror, and aligned to point to the center of the hood. 
 Markers on the car hood were used to initially align the camera for the first recording session and the camera was never moved from this position. These markers were left on the hood throughout the entire recording period for control.
 A polarization filter was used in some of the recordings to reduce windshield and hood glare.
 The camera was powered by and connected to a laptop computer through high speed USB 2.0.
 The raw data was read out using inilabs cAER software\footnote{\href{https://inilabs.com/support/software/caer/}{cAER support}} and streamed to the custom recording framework described in Sec.~\ref{sect:software} for further processing.

 \subsection{Vehicle control and diagnostic data}

Data was acquired using a Ford Mondeo MK 3 European Model. 
We used the OpenXC Ford Reference vehicle interface, that plugs into the 
passenger compartment OBDII port, to read out control and diagnostic data from the car's CAN bus. The vehicle interface connects to a host USB port\footnote{\href{http://openxcplatform.com/vehicle-interface/hardware.html}{OpenXC vehicle interface}}.

 The vehicle interface was programmed with the vendor-provided firmware for the Ford Mondeo MK 3 car model (``type 3'' firmware) and read out using the OpenXC python library.
 The raw data was passed to the custom recording software described in Sec.~\ref{sect:software}.
 The following quantities were read out at rates of about 10\,Hz each. 
 Likely targets for experiments in end-to-end learning are in boldface.

 \begin{itemize}
 \setlength{\parskip}{0pt}
    \setlength{\parsep}{0pt}  
    \setlength\itemsep{0em}
    \item \textbf{steering wheel angle} (degrees, up to 720${^\circ}$)
     \item \textbf{accelerator pedal position} (\% pressed),
     \item \textbf{brake pedal status} (pressed/not pressed),
     \item engine speed (rpm),
     \item vehicle speed (km/h),
     \item latitude,
     \item longitude,
     \item headlamp status (on/off),
     \item high beam status (on/off),
     \item windshield wiper status (on/off),
     \item odometer (km),
     \item torque at transmission,
     \item transmission gear position (gear no.),
     \item fuel consumed since restart,
     \item fuel level (\%),
     \item ignition status,
     \item parking brake status (on/off).
 \end{itemize}

 \subsection{Recording and viewing software}
 \label{sect:software}

 A python software framework
 \footnote{\href{https://code.ini.uzh.ch/jbinas/ddd17-utils}{ddd17-utils}} 
 for recording, viewing, and exporting the data was created for the main purpose of combining and synchronizing the data from the different input devices and storing it in a standardized file format.
 In particular, since the APS frames and DVS data are microsecond time-stamped on the camera using its own local clock, whereas the data provided by the vehicle interface is not, both data streams were augmented with the millisecond system time of the recording computer, which could then be used for synchronization. 
 With the vehicle interface streaming data at rates of only around 10\,Hz per recorded variable, such off-device time-stamping is justified.
 The computer time was  synchronized to a standard time server before recordings.
 The data was stored in HDF5 format, for which widely used libraries for various environments exist.
 Each data type (e.g. DVS events, steering wheel angle, vehicle speed...) was stored in a separate container, each containing one container for the system timestamp and one for the data.
 In this way, the system timestamp can be used for fast indexing and for synchronizing the data when reading.
 With data being provided at irregular intervals by the recording devices, each data type was stored in an event-driven fashion, such that different containers contain different numbers of samples. 
 The DAVIS data was stored in its native cAER AER-DAT3.1 format\footnote{\href{https://inilabs.com/support/software/fileformat/}{inilabs file formats}} in each HDF5 container.

\begin{figure}[ht]
     \centering
     \includegraphics[scale=0.3]{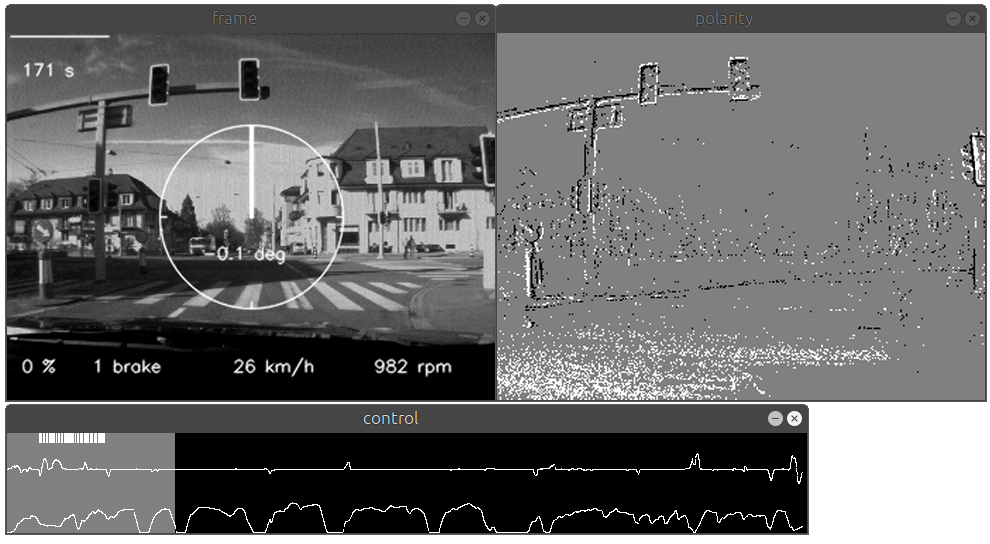}
     \caption{Example scenario visualized by the recording file viewer. The top panels show the DAVIS frames (left; overlaid with some driving data) and events (right), the bottom panel shows a progress bar as well as visualizations of different vehicle data (headlamp status at the top, steering angle in the middle, speed at the bottom).}
     \label{fig:viewer}
 \end{figure}

 In addition to the recording framework, a python-based viewer \texttt{view.py} visualizes the recorded DAVIS data together with selected vehicle data such as the steering angle or speed (Fig.~\ref{fig:viewer}).
The script \texttt{export.py} exports the data into frames for preparing data for further processing by machine learning algorithms.

\section{Recorded data}

\begin{figure}[ht]
\vskip 0.2in
\begin{center}
\centerline{\includegraphics[scale=0.8]{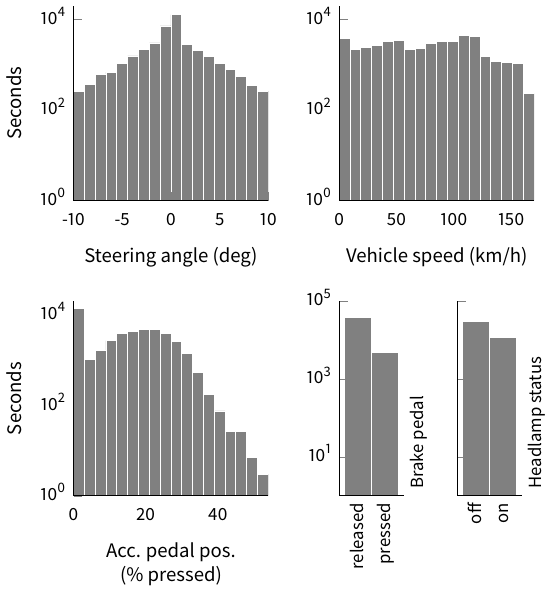}}
\caption{Statistical distribution of various recorded signals.}
\label{fig:stats}
\end{center}
\vskip -0.2in
\end{figure}

 In total, over 12\,h of data were recorded under various weather, driving, road, and lighting conditions on six consecutive days, covering over 1000\,km of different types of roads in Switzerland and Germany.
 Recordings were started and stopped manually and typically have durations of between a minute and an hour.
 The resulting recordings are summarized in Table~\ref{tab:summary}. 
 Fig.~\ref{fig:stats} shows the distributions of several recorded variables over the whole dataset. 
 Steering angles are dominated by straight driving and small deviations of $\pm 10{^\circ}$. 
 Speed is uniformly distributed over the range 0-160\,km/h.
 The automatically controlled headlight is on about half the time, indicating a substantial fraction of the data was captured in low-light conditions.

 %Implicit data includes
 %\begin{itemize}
 %    \item lighting conditions (through automatic headlamp status),
 %    \item wheather conditions (through automatic windshield wiper status).
 %\end{itemize}

 \begin{table}
 {\centering
 \begin{filecontents*}{recordings.csv}
file,description,condition,duration,size
%{1487326422},{mount calibration},,{91},{0.8}
%{1487329692},{test; parked in garage},,{10},{0.09}
%{1487333001},{test on campus (garage and outside)},{rain},{96},{1.2}
%{1487337800},{test on campus (garage and outside)},,{108},{1.1}
{1487339175},{cty},{wet},{347},{2.8}
%
{1487349453},{campus},{dark},{192},{1.7}
{1487350455},{fwy},{ngt, rain},{1404},{11.2}
{1487354030},{cty},{ngt, wet},{377},{3}
{1487354811},{cty},{ngt, wet},{190},{1.4}
{1487355025},{cty},{ngt, wet},{57},{0.4}
{1487355090},{cty, hwy},{ngt, wet},{984},{5.9}
{1487356509},{fwy},{ngt, wet},{2233},{12.4}
{1487417411},{fwy},{day},{2096},{18.2}
{1487419513},{fwy},{day},{1976},{18.3}
{1487424147},{m. fwy},{day},{3040},{30.3}
{1487427200},{fwy},{day},{1947},{17.6}
{1487430438},{fwy},{day},{3135},{26.2}
{1487433587},{fwy+cty},{ngt},{2355},{18.5}
{1487593224},{hwy},{day},{586},{5.3}
{1487594667},{fwy},{day},{2985},{29.7}
{1487597945},{cty},{evening},{50},{0.5}
{1487598202},{fwy},{day},{1882},{15.1}
{1487600962},{fwy},{day},{2143},{15.1}
{1487608147},{fwy},{evening},{1208},{9}
{1487609463},{fwy},{evening},{1458},{6.3}
%
{1487778564},{campus},{day},{101},{1.1}
{1487779465},{cty+hwy},{day},{1170},{11.2}
{1487781509},{campus},{evening},{127},{0.6}
{1487782014},{cty+hwy},{evening},{1118},{7.3}
%
{1487839456},{cty},{day, sun},{406},{5.7}
{1487842276},{cty},{day, sun},{625},{9.5}
{1487844247},{cty},{day, sun},{523},{7.5}
{1487846842},{twn+hwy},{day, sun},{1799},{20.6}
{1487849151},{twn},{day, sun},{429},{5.5}
{1487849663},{twn+hwy},{day, sun},{2863},{34.7}
{1487856408},{twn},{day, sun},{817},{13.2}
{1487857941},{twn},{day, sun},{99},{1.4}
{1487858093},{cty},{day, sun},{2421},{34.7}
{1487860613},{cty},{day, sun},{1065},{17.4}
{1487864316},{cty+fwy},{evening},{1087},{12.9}
\end{filecontents*}
\pgfplotstabletypeset[
     col sep=comma,
     string type,
     columns/file/.style={column name=File(.hdf5), column type={|l}},
     columns/description/.style={column name=Scene, column type={|l}},
     columns/condition/.style={column name=Cond., column type={|l}},
     columns/duration/.style={column name=T (s), column type={|l}},
     columns/size/.style={column name=GB, column type={|l|}},
     every head row/.style={before row=\hline,after row=\hline},
     every last row/.style={after row=\hline},
 ]{recordings.csv}
 \caption{Summary of the acquired data. Keys: hwy=highway, fwy=freeway, cty=city, twn=town, ngt=night. GB=size of recording in gigabytes. T=duration of recording in seconds.}
 \label{tab:summary}
 }
 \end{table}

\section{Experiments: Steering prediction network}

End-to-end learning of a control model is 
an attractive approach for self-driving applications, 
since it eliminates the need for tedious hand-labeling 
of the data or features -- a task which is prohibitive in the face of the enormous amounts of data acquired by 
today's vehicles \citep{bojarski2016end}.
The presented dataset has clear limitations, since it does not include other sensors such as LIDAR, does not include route information that would allow better prediction of user intentions, and the data tends to be unbalanced. 
Nevertheless, under certain conditions 
such as highway driving, driving along roads without 
turns onto other roads, or 
unpredictable user actions, it can be used to 
study the utility of of the data for prediction of measured user actions.

We trained simple steering prediction networks. These networks take input APS and/or DVS data and attempt to predict the instantaneous steering wheel angle. 
They are inspired by LeCun's early work \citep{lecun_off-road_2005}, the seminal open dataset from comma.ai \citep{santana2016commaai}, and by recent Nvidia~\citep{bojarski2016end} and unpublished VW studies.

 Our results compare the steering prediction accuracy of 
 networks operating on pure APS data to such operating on pure DVS data.
Our example implementation should be regarded as a preliminary study to validate the usability of the data and associated software. 
In particular, the experiments presented here are based on a small subset of the whole dataset (recordings 1487858093 and 1487433587 in Table~\ref{tab:summary}).
Work is ongoing to train more architectures using more of the data. 

\begin{figure}[ht]
\begin{center}
\centerline{\includegraphics[width=\columnwidth]{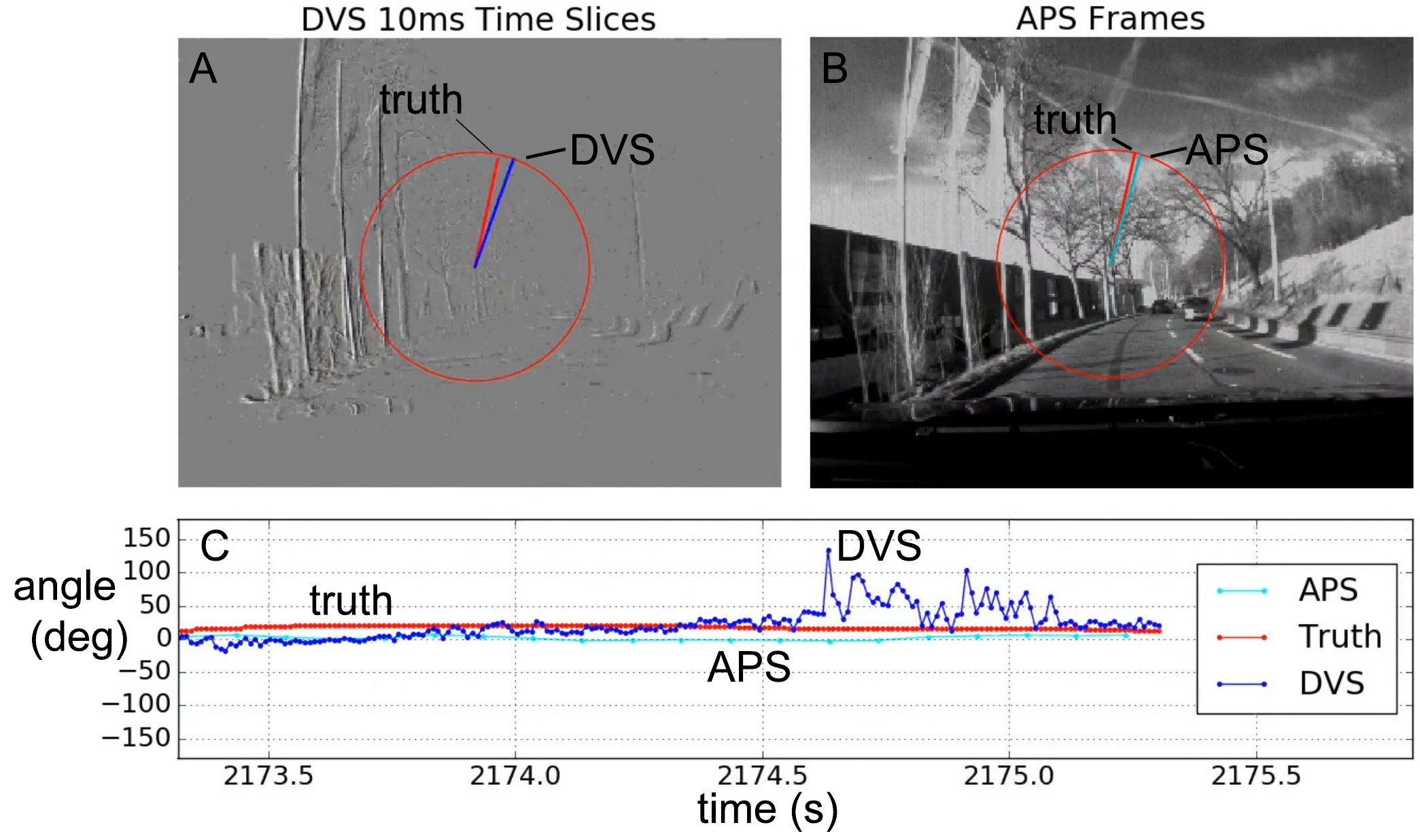}}
\caption{Steering prediction initial result. Comparison of our first APS and DVS steering prediction experiments. \textbf{A}: DVS frame and CNN output. \textbf{B}: APS frame and CNN output. \textbf{C}: segment of time history.}
\label{fig:cnninitial}
\end{center}
\end{figure}

Fig.~\ref{fig:cnninitial} shows our first results, obtained from a CNN with 4 convolutional layers, each with 8 feature maps and using 3x3 kernels, and trained on a single 1.5\,h recording. Each layer is followed by a 2x2 max pooling layer. The final feature map layer is mapped to a 64-unit fully connected (FC) layer. The FC layer is mapped to an output steering angle in the range $\pm 180{^\circ}$. The DVS and APS inputs were subsampled to 80x60 images. Input frame normalization was done as in \citet{moeys_steering_2016b}. 

Our quantitative accuracy results are too inconclusive to report but we have verified the usability of the dataset and tools. 
Further analysis is necessary and the subject of ongoing work.

\section{Conclusion}
The main result of this paper is to introduce the DDD17 first open dataset of DAVIS driving data with end-to-end labeling, along with necessary software tools. A preliminary study on an end-to-end steering angle prediction by a CNN show usability of the data.

\section*{Acknowledgements} 
{\small
We thank Dimitri Rettig and Anna Stockklauser for their help with recording some of the data, \href{https://www.inilabs.com}{inilabs} and the \href{http://sensors.ini.uzh.ch}{INI Sensors Group} for device support.
This work was made possible by funding from the EU projects \href{http://www.seebetter.eu/}{SeeBetter} and \href{http://www.visualise-project.eu/}{Visualise} and by Samsung.
}
 
\bibliography{bibliography}
\bibliographystyle{icml2017}

\end{document}